\documentclass[journal]{IEEEtran}

\ifCLASSINFOpdf
\else
\fi

\usepackage{amsmath}
\usepackage{graphicx}
\usepackage{float}
\usepackage{color}
\usepackage{amsthm}
\usepackage{MnSymbol}%
\usepackage{wasysym}%
\usepackage{pdfpages}
\usepackage{subfig}
\usepackage{hyperref}
\usepackage{cleveref}

\usepackage{algorithm}
\usepackage{algpseudocode}

\begin{document}
%
\title{Enhancing Federated Learning with Kolmogorov-Arnold Networks: A Comparative Study Across Diverse Aggregation Strategies}
%
%
%



\author{
    \IEEEauthorblockN{Yizhou~Ma\IEEEauthorrefmark{1}, Zhuoqin~Yang\IEEEauthorrefmark{2}, 
    Luis-Daniel~Ibáñez\IEEEauthorrefmark{1}}\\
    \IEEEauthorblockA{\IEEEauthorrefmark{1}School of Electronics and Computer Science,
    Faculty of Engineering and Physical Sciences,
    University of Southampton,
    Southampton, UK\\
    Email: \{ym7u22, l.d.ibanez\}@soton.ac.uk}\\
    \IEEEauthorblockA{\IEEEauthorrefmark{2}School of Computer Science, University of Nottingham,
    Ningbo, China\\
    Email: scxzy5@nottingham.edu.cn}\\

    \thanks{Corresponding author: Yizhou Ma.}
    \thanks{This preprint has not undergone peer review or any post-submission improvements or corrections. It was prepared prior to submission to, and has since been accepted at, ICIC 2025. The final Version of Record will be published in the ICIC 2025 proceedings by Springer.
 }
}

%
%


\maketitle

\begin{abstract}

Multilayer Perceptron (MLP), as a simple yet powerful model, continues to be widely used in classification and regression tasks. However, traditional MLPs often struggle to efficiently capture nonlinear relationships in load data when dealing with complex datasets. Kolmogorov-Arnold Networks (KAN), inspired by the Kolmogorov-Arnold representation theorem, have shown promising capabilities in modeling complex nonlinear relationships. In this study, we explore the performance of KANs within federated learning (FL) frameworks and compare them to traditional Multilayer Perceptrons. Our experiments, conducted across four diverse datasets demonstrate that KANs consistently outperform MLPs in terms of accuracy, stability, and convergence efficiency. KANs exhibit remarkable robustness under varying client numbers and non-IID data distributions, maintaining superior performance even as client heterogeneity increases. Notably, KANs require fewer communication rounds to converge compared to MLPs, highlighting their efficiency in FL scenarios. Additionally, we evaluate multiple parameter aggregation strategies, with trimmed mean and FedProx emerging as the most effective for optimizing KAN performance. These findings establish KANs as a robust and scalable alternative to MLPs for federated learning tasks, paving the way for their application in decentralized and privacy-preserving environments.

\end{abstract}


\IEEEpeerreviewmaketitle

\section{Introduction}
%
%
%
%
Multilayer Perceptron (MLP) \cite{gardner1998artificial}, as a simple yet powerful model, continues to be widely used in classification and regression tasks. However, traditional MLPs often struggle to efficiently capture nonlinear relationships \cite{danish2025kolmogorov} in load data when dealing with complex datasets. In contrast, the Kolmogorov-Arnold Network (KAN) \cite{liu2024kan}, based on the Kolmogorov-Arnold Representation Theorem, offers a promising alternative. KAN leverages the theorem to decompose complex multivariate functions into a linear combination of univariate functions, which are further fitted using learnable nonlinear transformation functions, demonstrating exceptional nonlinear approximation capabilities \cite{hou2024comprehensive}. Recently, several studies \cite{wang2024federated, sharma2024aggregation, hu2024heterogeneous, wu2024fedstem} on federated learning (FL) have been conducted, KAN also has been applied to a variety of tasks, including tabular data \cite{poeta2024benchmarking}, time series \cite{vaca2024kolmogorov}, and computer vision \cite{cheon2024demonstrating}, showcasing its robust performance. Nevertheless, while some efforts \cite{zeydan2024f, sasse2024evaluating} have been made to explore the application and performance of KAN in federated learning and other distributed scenarios, these studies remain incomplete and leave considerable room for further investigation.

In our research, we focus on exploring the performance of Kolmogorov-Arnold Networks within a federated learning framework compared to traditional Multilayer Perceptrons. We designed a series of experiments in a multi-client simulated environment. Both models were subjected to identical training and testing pipelines to ensure fairness and consistency in the evaluation process. The experiments covered both binary and multi-class classification tasks, providing a comprehensive assessment of the models' capabilities. Additionally, we evaluated a range of parameter aggregation strategies to identify the most effective methods for optimizing performance in federated learning scenarios. 

We identify three key research questions (Q1, Q2, Q3) in exploring the performance of Kolmogorov-Arnold Networks within a federated learning framework and propose feasible solutions. The first research question is (Q1): \textbf{How to simulate common federated learning tasks across diverse evaluation scenarios?} To address this challenge, we selected four widely-used datasets to simulate typical tasks in federated learning environments. These tasks encompass both binary and multi-class classification problems, ensuring a diverse range of evaluation scenarios for model performance. The second research question is (Q2): \textbf{How to evaluate the feasibility of KANs in federated learning and compare their performance with Multilayer Perceptrons?} To tackle this challenge, we designed a series of experiments in a multi-client simulated environment. Both KANs and MLPs were subjected to identical training and testing pipelines to ensure fairness and consistency. Performance comparisons were conducted across classification tasks, providing a comprehensive analysis of their strengths and weaknesses. The third research question is (Q3): \textbf{How to identify global parameter aggregation strategies that are better suited for KANs in federated learning?} To solve this challenge, we evaluated a variety of parameter aggregation strategies for each task. By comparing the performance of these strategies, we aimed to determine the most effective approach for optimizing KANs in federated learning and benchmarked the results against those of MLPs. This systematic exploration provides insights into how different aggregation methods impact the overall effectiveness and adaptability of KANs in distributed environments. Our major contributions are summarized as follows:
\begin{itemize}
    \item We present a comprehensive exploration of Kolmogorov-Arnold Networks in federated learning, providing a preliminary guideline for their broad applicability in diverse evaluation scenarios. This includes validating their robustness and adaptability across binary and multi-class classification tasks.

    \item We perform extensive evaluations to compare the performance of Kolmogorov-Arnold Networks with Multilayer Perceptrons under various FL configurations.

    \item We identify and thoroughly evaluate a range of parameter aggregation strategies to determine the most effective approaches for optimizing KAN performance in federated settings.

\end{itemize}

The remainder of this paper is organized as follows. Section \ref{sec:bc} provides a comprehensive review of related works. Section \ref{sec: market} details the proposed methodology, outlining the experimental setup, the integration of KANs into the federated learning framework, and the diverse aggregation strategies employed for global model updates. Section \ref{sec: empirical} presents the empirical evaluation, where we conduct extensive experiments across multiple datasets to assess the performance of KANs compared to traditional Multilayer Perceptrons under different federated learning configurations. Finally, Section \ref{sec:con} concludes the paper and presents the vision for future work.




\begin{figure*}[htbp]

		\centering
		\includegraphics[width=1.0\linewidth]{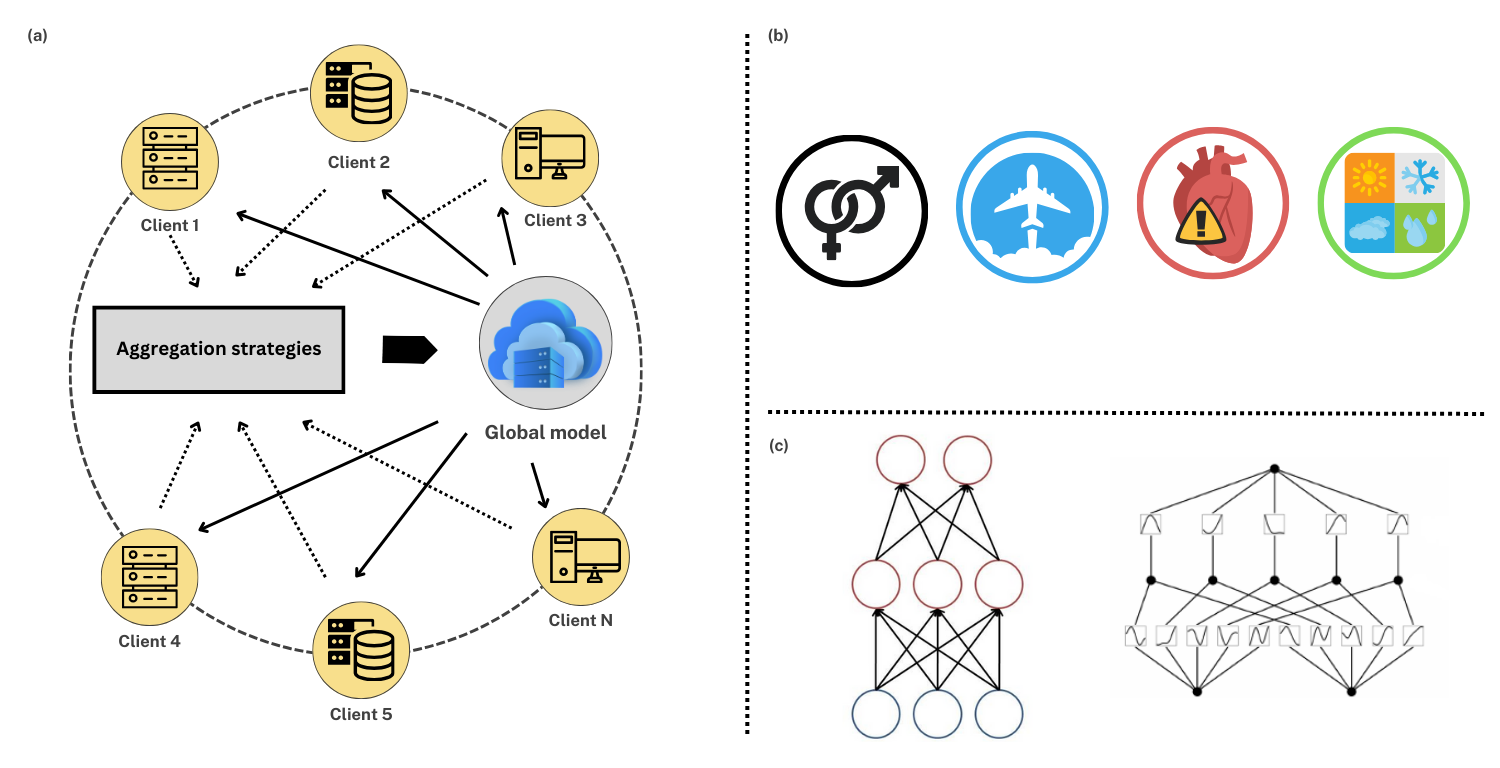}
		\caption{(a) The proposed framework for Federated Kolmogorov-Arnold Networks (F-KAN), demonstrating the interaction between clients, aggregation strategies, and the global model. The dotted and solid arrows represent bidirectional communication in the federated learning process. (b) Icons representing the four datasets used in this study: Gender Classification, Airline Passenger Satisfaction, Cardiovascular Disease Diagnosis, and Weather Type Classification. (c) Visualization of the two models compared in this research: a standard Multilayer Perceptron on the left and the Kolmogorov-Arnold Network on the right (adopted from \cite{liu2024kan}), highlighting their structural differences.}
		\label{fkan}
\end{figure*}

\section{Background}
\label{sec:bc}

Inspired by the Kolmogorov-Arnold Representation Theorem, Liu et al. \cite{liu2024kan} introduced Kolmogorov-Arnold Networks as a novel alternative to Multilayer Perceptrons. Unlike MLPs, which rely on linear weighted connections combined with fixed nonlinear activation functions, KANs replace linear weights with learnable univariate functions parameterized as splines on the edges. Several recent studies \cite{zeydan2024f, sasse2024evaluating,zeleke2024federated} have begun to explore the application and performance of KAN in federated learning. Zeydan et al.\cite{zeydan2024f} proposed Federated Kolmogorov-Arnold Networks (F-KAN), a novel integration of Kolmogorov-Arnold Networks within a federated learning framework. Through experiments conducted on the Iris dataset, they demonstrated that F-KANs significantly outperform Federated Multi-Layer Perceptrons (F-MLP) in terms of accuracy, precision, recall, F1-score, and stability. Specifically, F-KANs achieved 100\% accuracy on both the training and test datasets within just seven rounds of federated learning, showcasing their superior learning efficiency and robust performance compared to F-MLPs. However, their study is limited by the use of a single dataset for evaluation, which constrains the external validity of their findings. Further research is needed to validate Federated Kolmogorov-Arnold Networks across a broader range of datasets and classification tasks to fully assess their potential in practical decentralized environments. Sasse et al. \cite{sasse2024evaluating} evaluated Federated Kolmogorov-Arnold Networks in a more complex federated learning setting using the MNIST dataset. Their experiments compared two variations of F-KANs (Spline-KANs and RBF-KANs) against traditional Multilayer Perceptrons over 100 communication rounds in a highly non-iid partitioned environment with 100 clients. The study demonstrated that Spline-KANs consistently outperformed MLPs in terms of test accuracy across all communication rounds, particularly achieving significant gains in the early rounds, while RBF-KANs showed much lower performance, highlighting the sensitivity of KANs to activation function choices. Furthermore, Spline-KANs achieved similar accuracies to MLPs in half the training rounds, indicating faster convergence in federated settings. However, the evaluation was also restricted to a single dataset (MNIST). The authors also noted the need for more optimized aggregation algorithms for certain KAN architectures, such as RBF-KANs, to improve their applicability and performance. Zeleke et al. \cite{zeleke2024federated} investigates the integration of Kolmogorov-Arnold Networks into federated learning specifically for healthcare applications, leveraging ECG signal data for performance evaluation. Their work highlights the unique strength of KANs in capturing nonlinear time-series patterns through spline-based activation functions, which provide greater flexibility and efficiency compared to traditional MLPs. Using both real-world datasets, such as the MIT-BIH arrhythmia dataset, and synthetic ECG datasets, their results reveal that KANs significantly outperform MLPs in terms of accuracy and F1-score on the real dataset. Additionally, KANs demonstrate robust performance on synthetic data, confirming their generalizability across diverse datasets. Despite achieving superior predictive accuracy, their study acknowledges the computational overhead associated with KANs in federated learning environments, suggesting the need for optimization in resource-constrained settings. This research underscores the potential of KANs in privacy-preserving healthcare applications while calling attention to the challenges of computational efficiency and scalability in federated learning settings.

\section{Methodology}
\label{sec: market}


Figure 1 demonstrates the overall framework of our Federated Kolmogorov-Arnold Network implementation and the interactions between the global server and multiple clients. The process begins with the initialization of the global KAN model on the central server. Each client holds a unique portion of the dataset, ensuring that data privacy is preserved by keeping all local data decentralized. The clients independently train their local KAN models using their assigned data and subsequently transmit their model updates to the central server. Upon receiving these updates, the central server employs one of several aggregation strategies to update the global model. These aggregation strategies include average updates, median updates, trimmed mean updates, momentum updates, Nesterov momentum updates, Krum aggregation, and FedProx methods. This variety of strategies allows us to evaluate the robustness and adaptability of F-KANs under different optimization scenarios. The global model is then broadcast back to the clients, and this iterative process continues over a predefined number of communication rounds. The ultimate goal is to progressively refine the global model through collaborative learning while preserving data privacy. After the training rounds are completed, the updated global model is evaluated on an external test dataset to measure its performance across accuracy. Figure \ref{fkan} illustrates the overall workflow of our federated learning framework, depicting the interactions between clients, aggregation strategies, and the global model. To complement this, Algorithm \ref{al} provides the pseudo-code detailing the training and evaluation procedures for KAN and MLP models, covering different client configurations and aggregation strategies.

\begin{algorithm}
\caption{Federated Learning with \texttt{KAN/MLP}}
\label{al}
\begin{algorithmic}[0]
\small

\State \textbf{Given:} Number of clients $K$, Update method $M$
\State

\State \textbf{Global Model Initialization}
\State \quad $\text{global\_model} \gets \texttt{KAN/MLP}(\text{input\_size}, \text{output\_size})$
\State \quad $\text{global\_weights} \gets \textsc{ExtractParams}(\text{global\_model})$

\State

\State \textbf{Local Training Function} \quad \texttt{train\_on\_client}
\State \quad \textbf{Input:} (model, train\_loader, num\_epochs)
\State \quad \textbf{For each epoch} \textbf{in} $[1.. \text{num\_epochs}]$:
\State \quad \quad \textbf{For each batch} $(X, Y)$ \textbf{in} train\_loader:
\State \quad \quad \quad \textsc{Forward}(model, $X$)
\State \quad \quad \quad \textsc{ComputeLoss}($\text{outputs}, Y$) 
\State \quad \quad \quad \textsc{BackwardAndUpdate}(model.params)
\State \quad \textbf{return} \textsc{ExtractParams}(model)

\State

\State \textbf{Federated Rounds}
\State \quad \textbf{Initialize:} $\text{num\_rounds},\, \text{num\_epochs\_per\_round}$
\State \quad \textbf{For each} round \textbf{in} $[1.. \text{num\_rounds}]$:
\State \quad \quad $\text{client\_weights} \gets \emptyset$
\State \quad \quad \textbf{For each} client $c \textbf{ in } [1.. K]$:
\State \quad \quad \quad $\text{client\_model} \gets \textsc{Copy}(\text{global\_model})$
\State \quad \quad \quad $\text{trained\_weights} \gets \texttt{train\_on\_client}$
\State \quad \quad \quad \textsc{Append}($\text{client\_weights},\, \text{trained\_weights}$)

\State \quad \quad $\text{new\_weights} \gets \textsc{UpdateWeights}(\text{method} = M)$
\State \quad \quad \textsc{Assign}($\text{global\_model},\, \text{new\_weights}$)

\end{algorithmic}
\end{algorithm}

\begin{table*}[htbp]
\centering
\caption{Performance Comparison of KAN and MLP Across Four Datasets}
\renewcommand{\arraystretch}{1.2} 
\begin{tabular}{|l|cc|cc|cc|cc|}
\hline
\textbf{Dataset / Method}   & \multicolumn{2}{c|}{\textbf{3 Clients}} & \multicolumn{2}{c|}{\textbf{5 Clients}} & \multicolumn{2}{c|}{\textbf{10 Clients}} & \multicolumn{2}{c|}{\textbf{20 Clients}} \\ 
\cline{2-9}
                            & \textbf{KAN}      & \textbf{MLP}      & \textbf{KAN}      & \textbf{MLP}      & \textbf{KAN}      & \textbf{MLP}      & \textbf{KAN}      & \textbf{MLP}      \\ \hline
\multicolumn{9}{|c|}{\textbf{GENDER}} \\ \hline
Average                     & 96.94             & \textbf{96.94}             & \textbf{97.07}             & 96.80             & 96.80             & 96.67             & 96.67             & 96.67             \\
Median                      & 96.94             & 96.80             & 96.67             & 96.80             & 96.67             & 96.54             & 96.54             & 96.27             \\
Trimmed Mean                & 96.80             & 96.80             & 96.80             & 96.80             & 96.94             & 96.67             & \textbf{97.34}             & 96.54             \\
Momentum                    & \textbf{97.07}             & 96.80             & 95.87             & \textbf{97.20}             & 95.87             & \textbf{96.94}             & 92.14             & 95.07             \\
Nesterov Momentum           & 96.94             & 96.54             & 96.94             & 97.07             & 96.40             & 96.67             & 96.27             & 96.67             \\
Krum                        & 96.94             & \textbf{96.94}             & 96.94             & 96.54             & 96.80             & 96.40             & 96.80             & \textbf{96.80}             \\
FedProx                     & 96.94             & 96.80             & \textbf{97.07}             & 96.94             & \textbf{97.20}             & 96.80             & 96.40             & 96.67             \\ \hline
\multicolumn{9}{|c|}{\textbf{FLIGHT}} \\ \hline
Average                     & 96.51             & 95.71             & 96.50             & 95.68             & \textbf{96.47}             & \textbf{94.97}             & 96.20             & 94.13             \\
Median                      & 96.43             & 95.66             & 96.40             & 95.66             & 96.45             & 94.96             & 96.29             & \textbf{94.24}             \\
Trimmed Mean                & 96.45             & \textbf{96.05}             & 96.51             & \textbf{95.75}             & 96.33             & \textbf{94.97}             & 96.26             & 94.03             \\
Momentum                    & 96.15             & 94.12             & 96.01             & 93.80             & 95.40             & 91.24             & 94.92             & 90.13             \\
Nesterov Momentum           & 96.50             & 95.73             & \textbf{96.59}             & 95.60             & 96.45             & 94.89             & 96.23             & 94.11             \\
Krum                        & 96.02             & 95.82             & 95.94             & 95.65             & 96.00             & 94.34             & 95.65             & 93.86             \\
FedProx                     & \textbf{96.59}             & 95.82             & 96.41             & 95.73             & 96.42             & 94.80             & \textbf{96.27}             & 93.85             \\ \hline
\multicolumn{9}{|c|}{\textbf{CARDIO}} \\ \hline
Average                     & 73.22             & 73.23             & 73.11             & 73.02             & 73.18             & 72.76             & 73.06             & \textbf{72.62}             \\
Median                      & 73.29             & 73.16             & 73.22             & 72.92             & 73.33             & 72.53             & \textbf{73.24}             & 72.53             \\
Trimmed Mean                & 73.20             & 73.30             & 73.10             & 72.82             & 73.14             & \textbf{72.86}             & 73.27             & 72.57             \\
Momentum                    & \textbf{73.31}             & 72.57             & \textbf{73.28}             & 72.51             & 73.05             & 72.49             & 72.77             & 72.46             \\
Nesterov Momentum           & 73.26             & 73.14             & 73.18             & 72.99             & \textbf{73.36}             & 72.62             & 73.07             & 72.53             \\
Krum                        & 73.11             & 73.06             & 72.97             & 72.64             & 72.96             & 72.61             & 72.85             & 72.48             \\
FedProx                     & 73.27             & \textbf{73.36}             & 73.13             & \textbf{73.13}             & 73.13            & 72.62             & \textbf{73.24}             & \textbf{72.62}             \\ \hline
\multicolumn{9}{|c|}{\textbf{WEATHER}} \\ \hline
Average                     & 91.06             & \textbf{90.71}             & \textbf{91.16}             & 90.26             & 90.35             & 88.89             & 91.26             & 88.18             \\
Median                      & 91.21             & 90.51             & 91.11             & 90.35             & 91.06             & \textbf{91.11}             & 91.16             & \textbf{90.45}             \\
Trimmed Mean                & \textbf{91.41}             & 90.10             & 91.01             & 89.75             & \textbf{91.41}             & 89.24             & \textbf{91.46}             & 88.43             \\
Momentum                    & 91.06             & 90.76             & 90.30             & 90.00             & 89.85             & 88.94             & 89.34             & 83.79             \\
Nesterov Momentum           & 91.21             & \textbf{90.71}             & 90.66             & 90.71             & 91.16             & 89.04             & 91.31             & 87.58             \\
Krum                        & 90.20             & 90.20             & 90.91             & \textbf{90.81}             & 90.91             & 89.85             & 90.96             & 89.44             \\
FedProx                     & 91.11             & 90.61             & 90.76             & 90.71             & 90.91             & 90.10             & 91.26             & \textbf{90.45}             \\ \hline
\end{tabular}
\label{tab:comparison}
\end{table*}

\section{Empirical Evaluation}

\label{sec: empirical}

\subsection{Experiment setup}

Our study evaluates the performance of Kolmogorov-Arnold Networks in federated learning scenarios using four diverse datasets: Gender Classification \cite{kaggle_Gender}, Airline Passenger Satisfaction \cite{kaggle_Airline}, Cardiovascular Disease Diagnosis \cite{kaggle_Cardiovascular}, and Weather Type Classification \cite{kaggle_Weather}. The Gender Classification dataset consists of 5,000 samples, including individual facial feature data such as forehead length and lip thickness, and is used to predict user gender. The Airline Passenger Satisfaction dataset contains 129,880 samples covering various flight details and passenger information, with the goal of predicting passenger satisfaction levels. The Cardiovascular Disease Diagnosis dataset includes 70,000 samples spanning 11 features, such as age, blood pressure, and cholesterol levels, and aims to predict whether an individual has cardiovascular disease. Lastly, the Weather Type Classification dataset comprises 13,200 samples with daily weather and location information, with the objective of predicting weather types (e.g., sunny, cloudy, rainy, or snowy). These datasets encompass both binary and multi-class classification tasks, providing a broad spectrum of evaluation scenarios. The datasets were preprocessed with data cleaning, normalization, and random partitioning among clients to mimic real-world federated learning environments. To benchmark the effectiveness of KANs, we employed two types of neural network models, described as follows:

\begin{itemize}
    \item \textbf{KAN}: The primary model consists of four KAN layers with hidden dimensions set to [25, 50]. The output layer utilizes Sigmoid activation for binary classification tasks and Softmax activation for multi-class classification tasks.

    \item \textbf{MLP}: As a baseline, a fully connected neural network with the same depth and width as the KAN model was used for comparison. The MLP employs the classic ReLU activation function to achieve non-linear mappings.
    
\end{itemize}

Both models followed identical training and testing pipelines to ensure fairness in the experiments, with performance compared across binary and multi-class tasks. All experiments were conducted on a workstation equipped with an NVIDIA RTX 4090 GPU.

The experiments were conducted in federated learning environments with varying numbers of clients: 3, 5, 10, and 20. These scenarios simulate diverse federated learning setups, ranging from low to high client counts. In each scenario, the datasets were randomly partitioned across clients, with each client holding a slightly uneven number of samples to better reflect the real-world distribution of data. Each experiment included 20 and 60 communication rounds for KAN and MLP, respectively. During each round of communication, each client executed 3 local training epochs. The Adam optimizer was used with a learning rate of 0.005. To explore the suitability of KANs for federated learning, we incorporated a range of parameter aggregation strategies, including: Average Updates (FedAvg), Median Updates,Trimmed Mean Updates, Momentum Updates, Nesterov Momentum Updates, Krum Aggregation, FedProx Methods. By comparing these strategies across the datasets and tasks, we aimed to identify the most effective aggregation methods for optimizing KAN performance and benchmarked their results against those of MLPs.

\begin{figure*}[htbp] 
    \centering
    \includegraphics[width=0.49\textwidth]{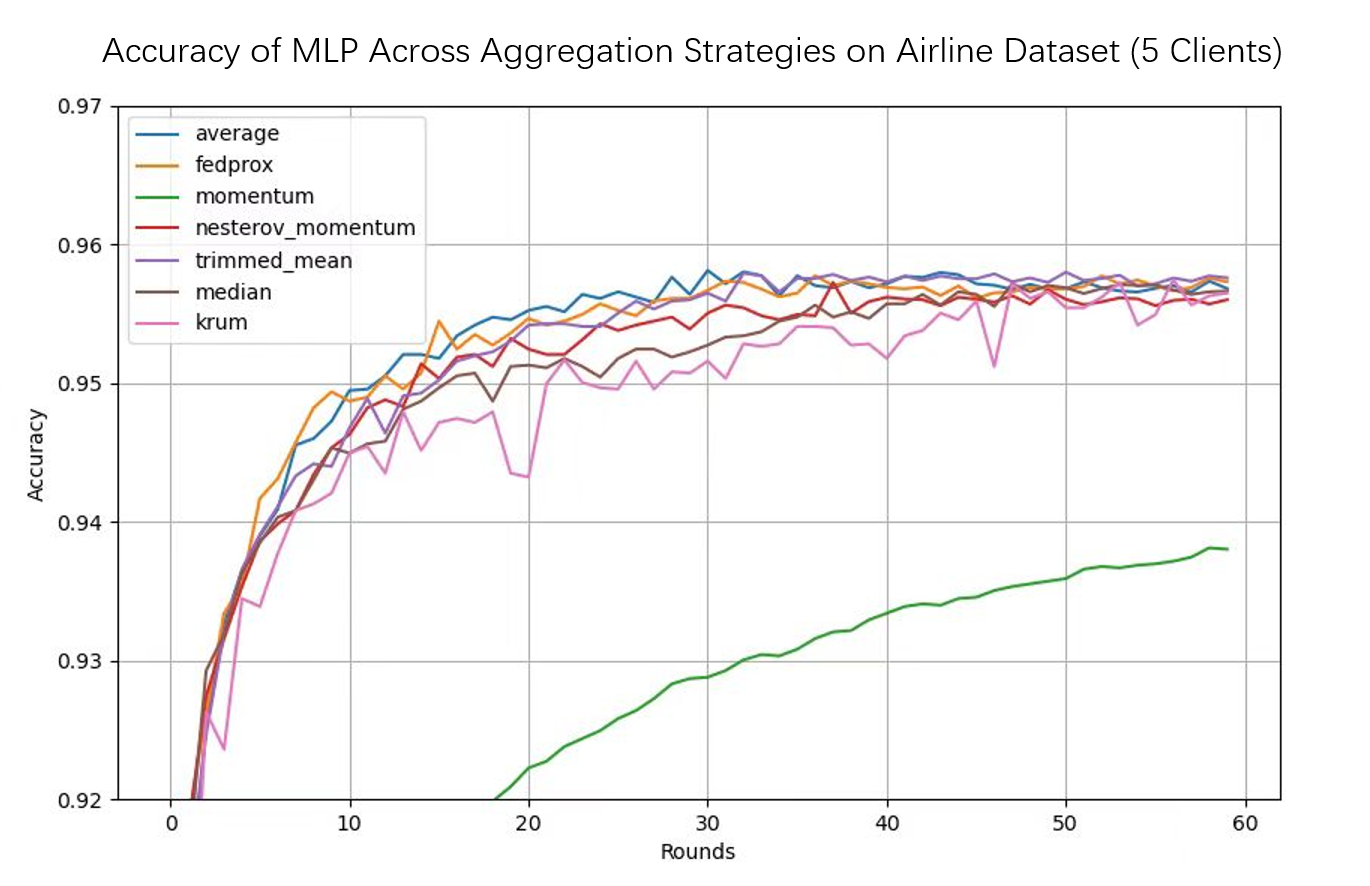} 
    \hspace{0.01\textwidth} 
    \includegraphics[width=0.48\textwidth]{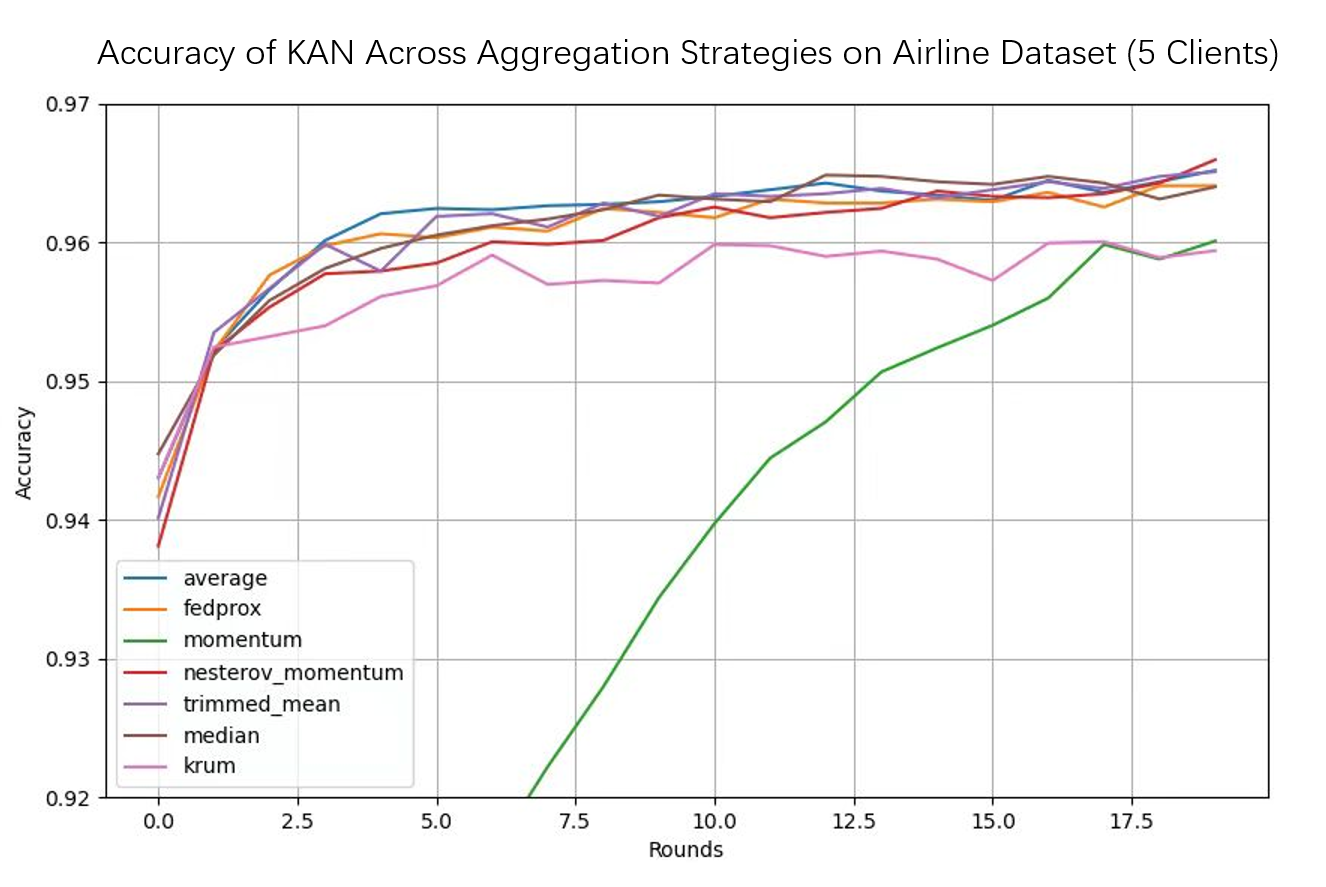} 
    \caption{(Left) Dataset Airline MLP with 5 clients. (Right) Dataset Airline KAN with 5 clients.}
    \label{fig:Dataset Airline}
\end{figure*}


	

\subsection{Empirical result}

\subsubsection{Gender Classification}

In this binary classification task, KANs demonstrated marginally better performance than MLPs across most aggregation strategies. Particularly with the trimmed mean and FedProx methods, KANs achieved a peak accuracy of 97.34\% in the 20-client scenario, outperforming MLPs, which achieved a maximum of 96.8\%.

\subsubsection{Airline Passenger Satisfaction}

Figure \ref{fig:Dataset Airline} presents the performance comparison of KAN and MLP models on the Airline dataset with 5 clients across 20 communication rounds. Both models were evaluated using multiple parameter aggregation strategies, including average, FedProx, momentum, Nesterov momentum, trimmed mean, median, and Krum.

The results highlight notable differences between KAN and MLP. KAN (Right) demonstrates faster convergence and higher stability, achieving an accuracy of approximately 96.6\% within the first 10 rounds, which remains consistent throughout the experiment. The trimmed mean and FedProx aggregation strategies yielded the best performance for KAN, maintaining high accuracy across rounds. In contrast, MLP (Left) converges more slowly, requiring more rounds to reach a similar accuracy level. While MLP achieved a maximum accuracy of approximately 95.7\%, slightly lower than KAN, it displayed greater variability in performance across different aggregation strategies. FedProx and average aggregation methods provided the most stable results for MLP, while momentum-based methods, such as Nesterov momentum, lagged in convergence.

This comparison underscores the advantages of KAN over MLP in federated learning scenarios, particularly in terms of convergence speed and robustness across aggregation strategies. These results further validate KAN’s suitability for tasks requiring efficient and consistent learning in distributed environments.

\subsubsection{Cardiovascular Disease Diagnosis}

For this dataset, the performance gap between KANs and MLPs was narrower. However, KANs still outperformed MLPs in most scenarios. Notably, with the trimmed mean strategy, KANs achieved an accuracy of 73.24\% in the 20-client configuration, demonstrating higher stability compared to MLPs.

\subsubsection{Weather Type Classification}

In this multi-class task, KANs significantly outperformed MLPs, especially in scenarios with fewer clients. Using the trimmed mean method, KANs achieved a peak accuracy of 91.41\% with 3 clients, compared to MLPs’ 90.10\%. Even in the 20-client scenario, KANs maintained a high accuracy of 91.26\%, whereas MLPs’ performance dropped to 88.18\%. These results highlight KANs’ ability to handle distributed, non-IID data more effectively in complex classification tasks.



\begin{figure}[htbp]
\centering
\includegraphics[width=0.50\textwidth]{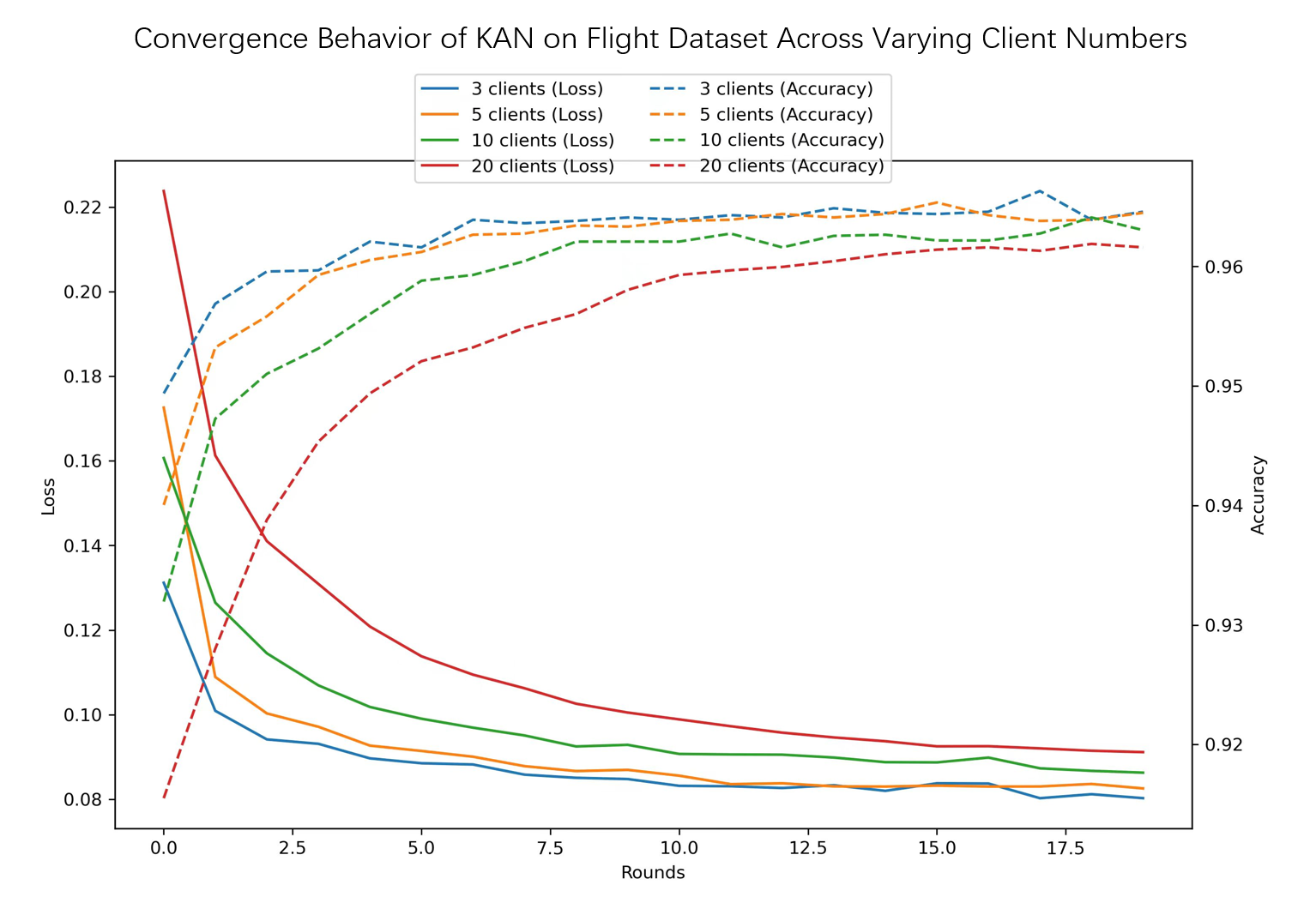}
\caption{Impact of Varying Client Numbers on Model Performance}
\label{Multiple clients}
\end{figure}

\begin{figure*}[htbp] 
    \centering
    \includegraphics[width=0.47\textwidth]{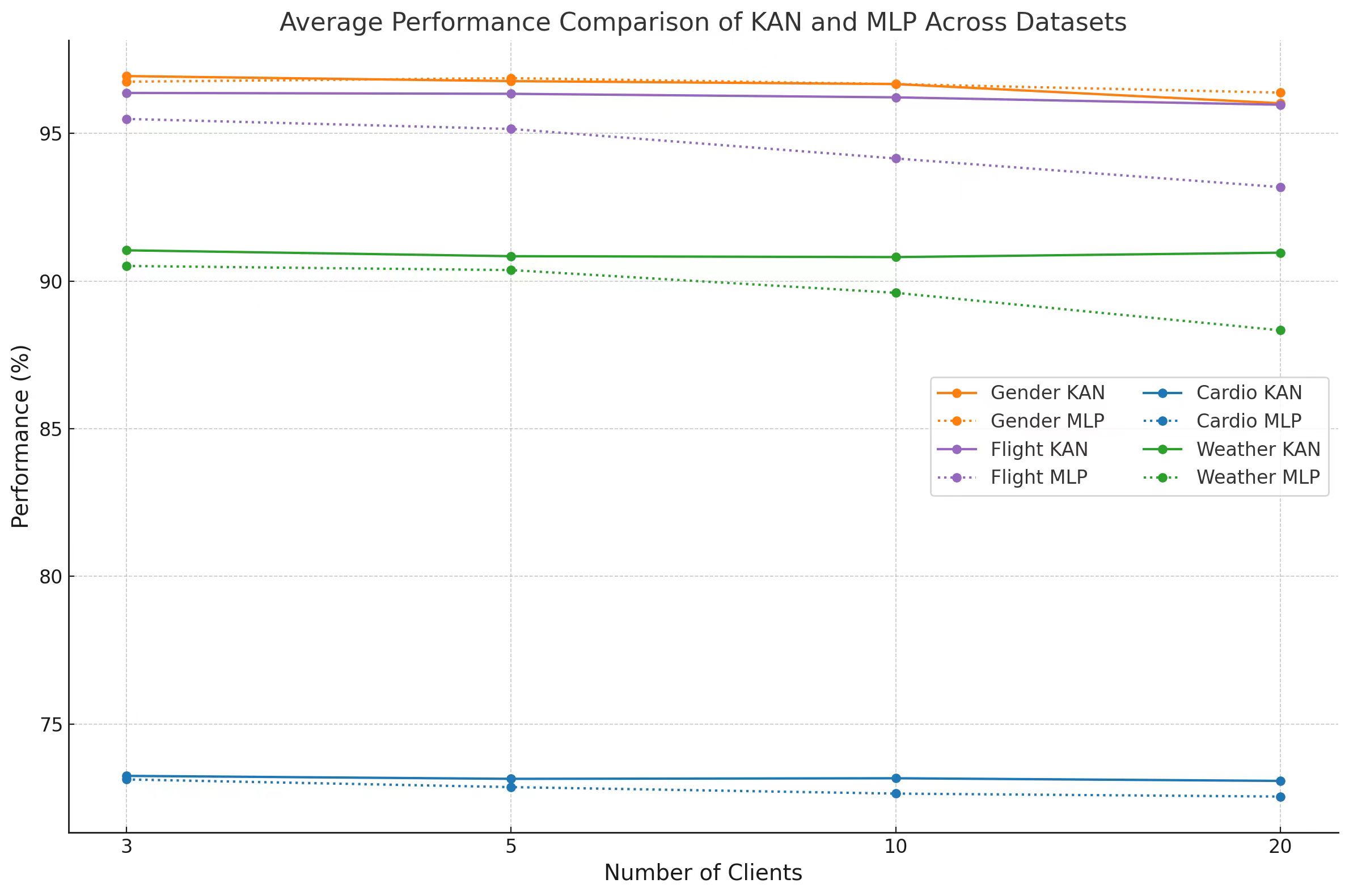} 
    \hspace{0.01\textwidth} 
    \includegraphics[width=0.47\textwidth]{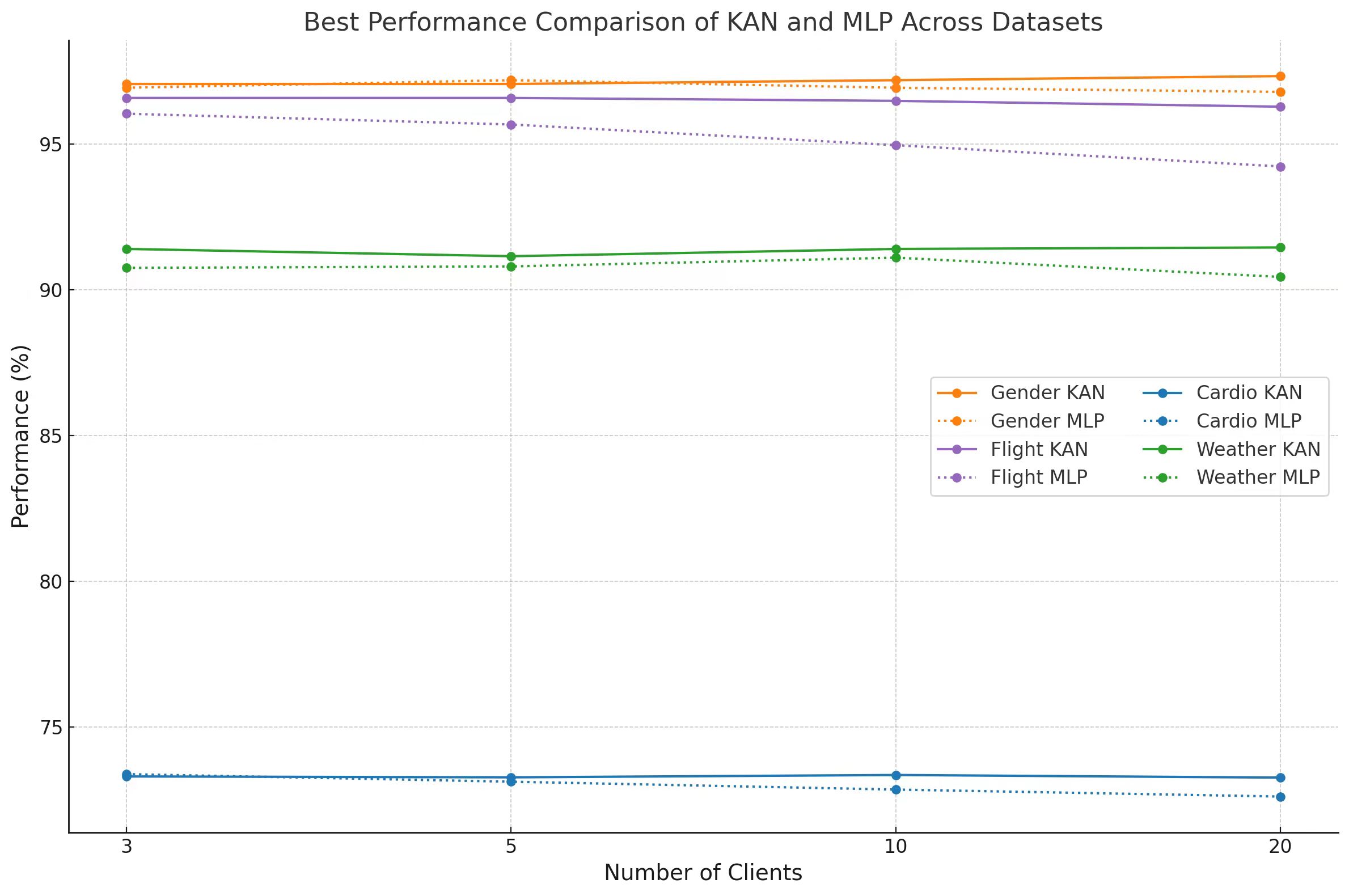} 
    \caption{Performance Comparison of KAN and MLP Across Different Datasets. The left figure presents the average performance of KAN and MLP across all aggregation strategies. The right figure highlights the best performance achieved by each model under the most optimal aggregation strategy}
    \label{fig: Performance}
\end{figure*}

Figure \ref{Multiple clients} illustrates the performance of the KAN model on the Flight dataset using the average aggregation method, evaluated across varying client numbers (3, 5, 10, and 20). The figure displays both the loss (solid lines) and accuracy (dashed lines) over 20 communication rounds, highlighting the model’s convergence behavior and performance consistency. The loss decreases steadily across all client configurations, with faster convergence observed in scenarios with fewer clients. For example, the 3-client setup demonstrates the quickest reduction in loss, reaching below 0.08 within 10 rounds, while the 20-client scenario converges more gradually due to greater heterogeneity in data distribution. Similarly, the accuracy trends align with the loss reduction patterns, where fewer clients enable faster and higher accuracy stabilization. The 3-client configuration achieves approximately 96\% accuracy within the first 10 rounds, while the 20-client configuration stabilizes around 92\% accuracy by the end of the 20 rounds. The higher variability in accuracy observed in the 20-client setup underscores the challenges posed by increased data distribution heterogeneity in federated learning environments. Overall, the figure demonstrates that the average aggregation method effectively supports the convergence and performance of KAN models, although the number of clients influences the speed and consistency of the results.

Figure \ref{fig: Performance} provides a comprehensive comparison of the performance of KAN and MLP across four datasets (Gender, Flight, Cardio, and Weather) under different client configurations (3, 5, 10, and 20). In Figure \ref{fig: Performance} (Left), the average performance of KAN across all aggregation methods consistently surpasses that of MLP. This trend is particularly evident as the number of clients increases, where KAN demonstrates greater robustness and stability, especially in datasets with higher data heterogeneity, such as Weather and Flight. MLP, on the other hand, exhibits a gradual decline in performance with increasing client numbers, indicating its sensitivity to non-IID data distributions. Figure \ref{fig: Performance} (Right) focuses on the best performance achieved by both models under their optimal aggregation methods. KAN not only matches but often exceeds MLP’s best performance, reinforcing its adaptability and effectiveness in federated learning scenarios. For instance, in the Cardio dataset, KAN achieves approximately 73\% accuracy even with 20 clients, while MLP shows a slight decrease in accuracy under similar conditions. These results collectively underline KAN’s superior ability to handle federated learning challenges, including client heterogeneity and data imbalance, making it a more reliable and scalable choice compared to MLP.

The empirical results demonstrate that KANs consistently outperform MLPs across most tasks and aggregation strategies, particularly under the trimmed mean and FedProx methods. Notably, KANs exhibited greater stability as the number of clients increased, a crucial property for federated learning environments with highly distributed data. In contrast, MLPs showed a significant decline in accuracy as the number of clients increased, highlighting their sensitivity to non-IID data distributions and client heterogeneity. Moreover, KANs required considerably fewer communication rounds to converge compared to MLPs, showcasing their efficiency in federated learning scenarios. Additionally, sensitivity to learning rate adjustments was observed under momentum-based strategies, suggesting that fine-tuning this parameter could further enhance KAN performance. In summary, KANs’ robustness, coupled with their superior accuracy, stability, and faster convergence in diverse federated learning configurations, underscores their potential as a more effective and scalable alternative to traditional MLPs in decentralized data scenarios.

\section{Conclusion and Future Work}
\label{sec:con}

In this paper, we presented an in-depth exploration of Kolmogorov-Arnold Networks (KANs) within federated learning frameworks and compared their performance to traditional Multilayer Perceptrons (MLPs). Through extensive experiments across four diverse datasets, we demonstrated that KANs consistently outperformed MLPs in terms of accuracy, stability, and convergence efficiency. KANs showed remarkable robustness against the challenges posed by increasing client numbers and non-IID data distributions, achieving superior performance across both average and best aggregation strategies. In particular, KANs required fewer communication rounds to converge compared to MLPs, making them more efficient in federated learning environments. Furthermore, KANs exhibited less sensitivity to increasing client numbers, maintaining stable performance even under high client heterogeneity, whereas MLPs experienced notable accuracy degradation.

While our study focused on tabular datasets, the performance of KANs on other data types, such as sequential, image, or graph-based data, remains unexplored. Extending our analysis to these domains could provide further insights into KANs’ applicability and versatility. Additionally, the parameter aggregation strategies evaluated in this study were designed for general federated learning models. Future research could develop aggregation strategies tailored specifically for KANs, leveraging their unique structural properties to enhance performance and adaptability in decentralized environments.

Our study also evaluated multiple parameter aggregation strategies, including FedProx, trimmed mean, and momentum-based methods, identifying effective approaches for optimizing KAN performance. Notably, the trimmed mean and FedProx strategies consistently yielded the best results, further demonstrating KANs’ adaptability to different federated learning configurations.



%





\ifCLASSOPTIONcaptionsoff
  \newpage
\fi



%

\bibliographystyle{IEEEtran}
\bibliography{bibtex/IEEEabrv}

\begin{thebibliography}{10}
\providecommand{\url}[1]{#1}
\csname url@samestyle\endcsname
\providecommand{\newblock}{\relax}
\providecommand{\bibinfo}[2]{#2}
\providecommand{\BIBentrySTDinterwordspacing}{\spaceskip=0pt\relax}
\providecommand{\BIBentryALTinterwordstretchfactor}{4}
\providecommand{\BIBentryALTinterwordspacing}{\spaceskip=\fontdimen2\font plus
\BIBentryALTinterwordstretchfactor\fontdimen3\font minus \fontdimen4\font\relax}
\providecommand{\BIBforeignlanguage}[2]{{%
\expandafter\ifx\csname l@#1\endcsname\relax
\typeout{** WARNING: IEEEtran.bst: No hyphenation pattern has been}%
\typeout{** loaded for the language `#1'. Using the pattern for}%
\typeout{** the default language instead.}%
\else
\language=\csname l@#1\endcsname
\fi
#2}}
\providecommand{\BIBdecl}{\relax}
\BIBdecl

\bibitem{gardner1998artificial}
M.~W. Gardner and S.~Dorling, ``Artificial neural networks (the multilayer perceptron)—a review of applications in the atmospheric sciences,'' \emph{Atmospheric environment}, vol.~32, no. 14-15, pp. 2627--2636, 1998.

\bibitem{danish2025kolmogorov}
M.~U. Danish and K.~Grolinger, ``Kolmogorov--arnold recurrent network for short term load forecasting across diverse consumers,'' \emph{Energy Reports}, vol.~13, pp. 713--727, 2025.

\bibitem{liu2024kan}
Z.~Liu, Y.~Wang, S.~Vaidya, F.~Ruehle, J.~Halverson, M.~Solja{\v{c}}i{\'c}, T.~Y. Hou, and M.~Tegmark, ``Kan: Kolmogorov-arnold networks,'' \emph{arXiv preprint arXiv:2404.19756}, 2024.

\bibitem{hou2024comprehensive}
Y.~Hou and D.~Zhang, ``A comprehensive survey on kolmogorov arnold networks (kan),'' \emph{arXiv preprint arXiv:2407.11075}, 2024.

\bibitem{wang2024federated}
H.~Wang, R.~Li, Z.~Xu, J.~Li, I.~King, and J.~Liu, ``Federated learning for vehicle trajectory prediction: Methodology and benchmark study,'' in \emph{2024 International Joint Conference on Neural Networks (IJCNN)}.\hskip 1em plus 0.5em minus 0.4em\relax IEEE, 2024, pp. 1--8.

\bibitem{sharma2024aggregation}
N.~Sharma, M.~Raj, and D.~Mishra, ``Aggregation-assisted proxyless distillation: A novel approach for handling system heterogeneity in federated learning,'' in \emph{2024 International Joint Conference on Neural Networks (IJCNN)}.\hskip 1em plus 0.5em minus 0.4em\relax IEEE, 2024, pp. 1--9.

\bibitem{hu2024heterogeneous}
Y.~Hu, R.~Liu, J.~Zhang, Z.-A. Huang, L.~Song, and K.~C. Tan, ``Heterogeneous structured federated learning with graph convolutional aggregation for mri-based mental disorder diagnosis,'' in \emph{2024 International Joint Conference on Neural Networks (IJCNN)}.\hskip 1em plus 0.5em minus 0.4em\relax IEEE, 2024, pp. 1--8.

\bibitem{wu2024fedstem}
D.~Wu, S.~Pateria, B.~Subagdja, and A.-H. Tan, ``Fedstem-adl: A federated spatial-temporal episodic memory model for adl prediction,'' in \emph{2024 International Joint Conference on Neural Networks (IJCNN)}.\hskip 1em plus 0.5em minus 0.4em\relax IEEE, 2024, pp. 1--8.

\bibitem{poeta2024benchmarking}
E.~Poeta, F.~Giobergia, E.~Pastor, T.~Cerquitelli, and E.~Baralis, ``A benchmarking study of kolmogorov-arnold networks on tabular data,'' \emph{arXiv preprint arXiv:2406.14529}, 2024.

\bibitem{vaca2024kolmogorov}
C.~J. Vaca-Rubio, L.~Blanco, R.~Pereira, and M.~Caus, ``Kolmogorov-arnold networks (kans) for time series analysis,'' \emph{arXiv preprint arXiv:2405.08790}, 2024.

\bibitem{cheon2024demonstrating}
M.~Cheon, ``Demonstrating the efficacy of kolmogorov-arnold networks in vision tasks,'' \emph{arXiv preprint arXiv:2406.14916}, 2024.

\bibitem{zeydan2024f}
E.~Zeydan, C.~J. Vaca-Rubio, L.~Blanco, R.~Pereira, M.~Caus, and A.~Aydeger, ``F-kans: Federated kolmogorov-arnold networks,'' \emph{arXiv preprint arXiv:2407.20100}, 2024.

\bibitem{sasse2024evaluating}
A.~M. Sasse and C.~M. de~Farias, ``Evaluating federated kolmogorov-arnold networks on non-iid data,'' \emph{arXiv preprint arXiv:2410.08961}, 2024.

\bibitem{zeleke2024federated}
S.~N. Zeleke and M.~Bochicchio, ``Federated kolmogorov-arnold networks for health data analysis: A study using ecg signal,'' in \emph{2024 IEEE International Conference on Big Data (BigData)}.\hskip 1em plus 0.5em minus 0.4em\relax IEEE, 2024, pp. 8070--8077.

\bibitem{kaggle_Gender}
\BIBentryALTinterwordspacing
Kaggle, ``Gender classification dataset,'' 2021. [Online]. Available: \url{https://www.kaggle.com/datasets/elakiricoder/gender-classification-dataset}
\BIBentrySTDinterwordspacing

\bibitem{kaggle_Airline}
\BIBentryALTinterwordspacing
T.~K, ``Airline passenger satisfaction,'' 2020. [Online]. Available: \url{https://www.kaggle.com/datasets/teejmahal20/airline-passenger-satisfaction}
\BIBentrySTDinterwordspacing

\bibitem{kaggle_Cardiovascular}
\BIBentryALTinterwordspacing
S.~U, ``Cardiovascular disease dataset,'' 2019. [Online]. Available: \url{https://www.kaggle.com/datasets/sulianova/cardiovascular-disease-dataset}
\BIBentrySTDinterwordspacing

\bibitem{kaggle_Weather}
\BIBentryALTinterwordspacing
Kaggle, ``Weather type classification,'' 2024. [Online]. Available: \url{https://www.kaggle.com/datasets/nikhil7280/weather-type-classification}
\BIBentrySTDinterwordspacing

\end{thebibliography}

%








\end{document}